\documentclass[sigconf,nonacm]{acmart}
\AtBeginDocument{%
  }

\usepackage{xspace}
\usepackage{dsfont}

\newcommand{\algoName}{\textsc{GasRL}\xspace}

\usepackage{lipsum}

\begin{document}

\title{Natural-gas storage modelling \\ by deep reinforcement learning}

% CONTRIBUTI 
% 1. modello/environment di RL
% 2. selezione di algoritmi
% 3. analisi del profile regolamentare

\author{Tiziano Balaconi}
\affiliation{%
  \institution{Università Roma Tre}
  \country{Italy}
  }
  \authornote{Work done during an internship at Banca d'Italia$^{\ddag}$.}
\orcid{0009-0005-9275-7178}

\author{Aldo Glielmo}
\affiliation{%
\institution{Banca d'Italia$^{\ddag}$}
  \country{Italy}
  }
  \authornote{aldo.glielmo@bancaditalia.it, marco.taboga@bancaditalia.it. \\ $^\ddag$The views and opinions expressed in this paper are those of the authors and do not necessarily reflect the official policy or position of Banca d’Italia.\\
  This article was published in the \emph{Proceedings of the 6th ACM International Conference on AI in Finance}, and is also available at \url{https://doi.org/10.1145/3768292.3770348}.}
\orcid{0000-0002-4737-2878} 

\author{Marco Taboga}
\affiliation{%
\institution{Banca d'Italia$^{\ddag}$}
  \country{Italy}
  }
\authornotemark[2]
\orcid{0000-0002-5611-5910}

\renewcommand{\shortauthors}{Bacaloni, Glielmo, Taboga}

\begin{abstract}
We introduce \algoName, a simulator that couples a calibrated representation of the natural gas market with a model of storage-operator policies trained with deep reinforcement learning (RL). 
We use it to analyse how optimal stockpile management affects equilibrium prices and the dynamics of demand and supply.
We test various RL algorithms and find that Soft Actor Critic (SAC) exhibits superior performance in the \algoName environment: multiple objectives of storage operators -- including profitability, robust market clearing and price stabilisation -- are successfully achieved.
Moreover, the equilibrium price dynamics induced by SAC-derived optimal policies have characteristics, such as volatility and seasonality, that closely match those of real-world prices. Remarkably, this adherence to the historical distribution of prices is obtained without explicitly calibrating the model to price data. 
We show how the simulator can be used to assess the effects of EU-mandated minimum storage thresholds. We find that such thresholds have a positive effect on market resilience against unanticipated shifts in the distribution of supply shocks. For example, with unusually large shocks, market disruptions are averted more often if a threshold is in place.

%
%Accurate models of gas storage operators are important for understanding the dynamics of natural gas markets and the impacts of regulatory interventions, such as the recent EU regulation on gas storage stockpiles.
% To rigorously evaluate EU gas‐storage regulations—specifically the mandate to fill at least 90\% of available capacity— the present analysis introduces a simulation‐based framework driven by a reinforcement‐learning (RL) “market economy” agent. 
% %
% At each discrete time step, a monopolistic RL agent simultaneously sets purchase and sale prices and chooses injection and withdrawal volumes to maximize cumulative financial return. 
% %
% We examine two configurations: (1) a baseline without any explicit inventory constraint; and (2) an augmented model that imposes an annual penalty if the agent’s storage level on November 1st falls below 90\% of capacity. Under the unconstrained baseline, the agent still behaves rationally, maintaining inventories just below 75\% despite no regulatory requirement. 
% %
% When the fill‐level constraint is enforced, the agent exceeds the mandated threshold on average at the start of November and reduces compliance violations to near zero—albeit at the expense of somewhat lower overall profitability.
\end{abstract}




\maketitle

%TODO: [CONSIDERAZIONI DA METTERE NELLA PARTE DEL SIMULATORE FORSE]The training phase permette all'agente di apprendere la strategic migliore sulla base del trade-off, intrinseco ad ogni algoritmo di RL, tra exploitation ( delle cononoscenze  già consolidate) e exploration ( di azioni alternative): al crescere del numero di step di allenamento, la reward ottenuta in fase di test cresce costantemente fino a convergenza, visto l'apprendimento costante da  

\section{Introduction}

% \para{ the policy problem: the EU regulation on gas storage }

In June 2022, the European Union required Member States to ensure that their underground natural gas storage facilities are at least 90\% full by the 1st of November of each year (80\% for the transitional year 2022) \cite{EU2022_1032}.
In July 2025, the refilling obligation was made slightly more flexible: the target can now be met anytime between the 1st of October and the 1st of December. This change marked yet another step in a long evolution of gas-storage regulation that started more than two decades earlier \cite{EU1998_30, EU2003_55, EU2009_715, EU2017_1938}.
The 90\% refilling target has proved controversial, with several governments and market participants arguing that a rigid 90\% target can distort seasonal prices, inflate summer wholesale prices and place a disproportionate financial burden on countries with large storage capacities \cite{Abnett2025a,Jack2025}. Although recent negotiations to make the regulation less stringent \cite{Abnett2025c} did not lead to changes in the target, the debate around its adequacy and effectiveness remains open, as it involves a complex assessment of the trade-off between market efficiency and energy security \cite{Reuters2025}.

% \para{ our approach: simulation-based testing of scenarios with a novel RL market economy }

In this work, we provide a model of the natural gas market, which we use to simulate and analyse the decision-making process of a gas storage operator and its consequences for the dynamics of natural gas prices and stockpiles.
This simulator can help policymakers and market participants better understand and analyse the effects of regulations, such as the EU regulation of June 2022, as well as the impact of specific market shocks.
We construct the simulator, which we dub `\algoName', in two steps.
First, we set up an environment that reproduces the main characteristics of the Italian gas market, one of the largest gas markets in the EU.
Then, we use state-of-the-art deep reinforcement learning (RL) methods to model the optimal interactions of a monopolistic storage operator with the environment (the assumption of a single operator is arguably realistic, as the Italian energy infrastructure company SNAM owns around 90\% of the country's storage capacity \cite{IEA_Italy_EPR_2023}).

We train several storage agents, with different RL algorithms and sets of hyperparameters. 
We find that agents trained with the Soft Actor Critic (SAC) algorithm perform better than the others in the \algoName environment, and that their decision policy gives rise to realistic market dynamics.
Notably, we find that the simulator accurately reproduces the volatility and seasonality of real-world prices without having been explicitly trained to match them.
We conclude by demonstrating the practical utility of the simulator through an analysis of how refilling targets affect prices, profitability, and market stability.

\subsection*{Related work.}
\vspace{0.2cm}
\noindent
\textbf{RL to model economic agents.}
The use of RL to train rational optimising economic agents within simulation settings has recently seen a surge of interest.
Its adoption began predominantly in the finance sector \cite{dixon2020machine}, where it has been applied to trading \cite{karpe2020multi,ardon2021towards} — including market making \cite{mascioli2024financial} and hedging strategies \cite{gao2023deeper} — and it has spurred the development of specialised open-source software \cite{amrouni2021abides,phantom2023}.
RL applications are also found in macroeconomic analysis \cite{atashbar2022deep}, where the methodology has been used to extend traditional general equilibrium models \cite{MiEtAl2023, dwarakanath2024tax,gabriele2025heterogeneous} or the capabilities of agent-based models \cite{glielmo2023reinforcement,brusatin2024simulating,adage2025,agrawal2025robust}.

\vspace{0.2cm}
\noindent
\textbf{RL in energy-systems simulations.}
While the applications of RL schemes to model the behaviour of economic agents in energy markets remain comparatively more limited, several existing papers can be connected with the present work.
For example, in \cite{di2025reinforcement} an RL agent trained with the Deep Deterministic Policy Gradient (DDPG) algorithm learns how to submit continuous offering curves in the European day-ahead electricity market. By optimally adjusting supply offers to market conditions, the agent significantly improves long-term profits, as compared to bidders implementing static strategies.
In \cite{harder2023finding}, the authors develop a multi-agent Twin Delayed Deep Deterministic Policy Gradient (TD3) framework to model groups of hydro-storage units in the German wholesale electricity market. They demonstrate that individually trained RL agents bidding in a decentralised yet competition-aware manner can accurately replicate real-world dispatch behaviours.
The work perhaps most closely related to the present one is \cite{curin2021deep}, where a deep learning framework inspired by, but not strictly based on RL, is used to optimise underground natural gas storage operations.

\vspace{0.2cm}
\noindent
\textbf{EU gas-storage modelling.}
EU gas storage and its regulation have been studied in a number of research papers and technical reports.
In \cite{FernandezBlanco2023}, the authors propose a partial-equilibrium model of the EU gas market to assess the impact of storage obligations. 
Specifically, they use the METIS simulator \cite{sakellaris2018metis} and analyse the evolution of market conditions under different combinations of uncertain input parameters. 
%The model runs separately for each sample, simulating system behavior under different scenarios. generating diverse combinations of uncertain input parameters. The model runs separately for each sample, simulating system behavior under different scenarios.  
%
In \cite{McWilliams2023}, a scenario-based assessment framework is used, together with estimates of the elasticity of daily gas demand to temperatures, to derive different demand profiles compatible with the achievement of storage targets during the 2023 energy crisis.
%
%For each scenario, they calculate required demand reductions—as percentage cuts relative to the 5‑year average—to meet a 90\% storage target by October 1st, 2023. This methodology blends supply assumptions with demand‑driven scenario analysis to inform policy planning.
%
In \cite{ENTSOG2022}, the authors generate scenarios for a single refilling season (from April to September); they simulate storage injections using PLEXOS, a commercial software tool for energy-systems modelling.
%
%In \cite{VIS2024}, the authors perform an empirical classification of national gas storage measures into groups such as mandatory stockholding, designation of entities tasked with refilling mandates, and voluntary filling schemes.
%The approach includes assessing the impact of these measures on market functioning and supply security.
%
In \cite{ECA2024}, the authors select a representative sample of EU Member States based on their gas supply profiles and crisis exposure and perform a comparative benchmarking.

\vspace{0.2cm}
\noindent
\textbf{Demand and supply on gas markets.}
The proposed specification of the \algoName environment relies on the vast literature that describes and analyses the functioning of natural gas markets (see, e.g., \cite{emiliozzi2025european} and the references therein).
In particular, our specification of the demand and supply equations draws from the analyses carried out by \cite{emiliozzi2025unveiling}, who use panel local projections to estimate impulse response functions showing how Italian gas consumption reacts over time and across economic sectors to unexpected supply shocks. 
They provide evidence that the stickiness of demand plays a crucial role in shaping market responses to shocks, a fact that we explicitly take into account in designing our simulator. 

\vspace{0.5cm}
The rest of this work is structured as follows. 
In Sec. \ref{sec:simulator} we describe the \algoName simulator, dividing the discussion between the RL environment and the RL agent.
In Sec. \ref{sec:experimental_setup} we provide details about the parametrisation of the environment and the training and testing procedures we followed to carry out the simulations.
In Sec. \ref{sec:results} we illustrate the results of our experiments, showcasing the simulator's realism, its adherence to real-world data, and its suitability as an instrument of policy analysis.
Finally, in Sec. \ref{sec:conclusions} we conclude.

\begin{figure}[t]
    \centering
    \includegraphics[width=1.0\linewidth]{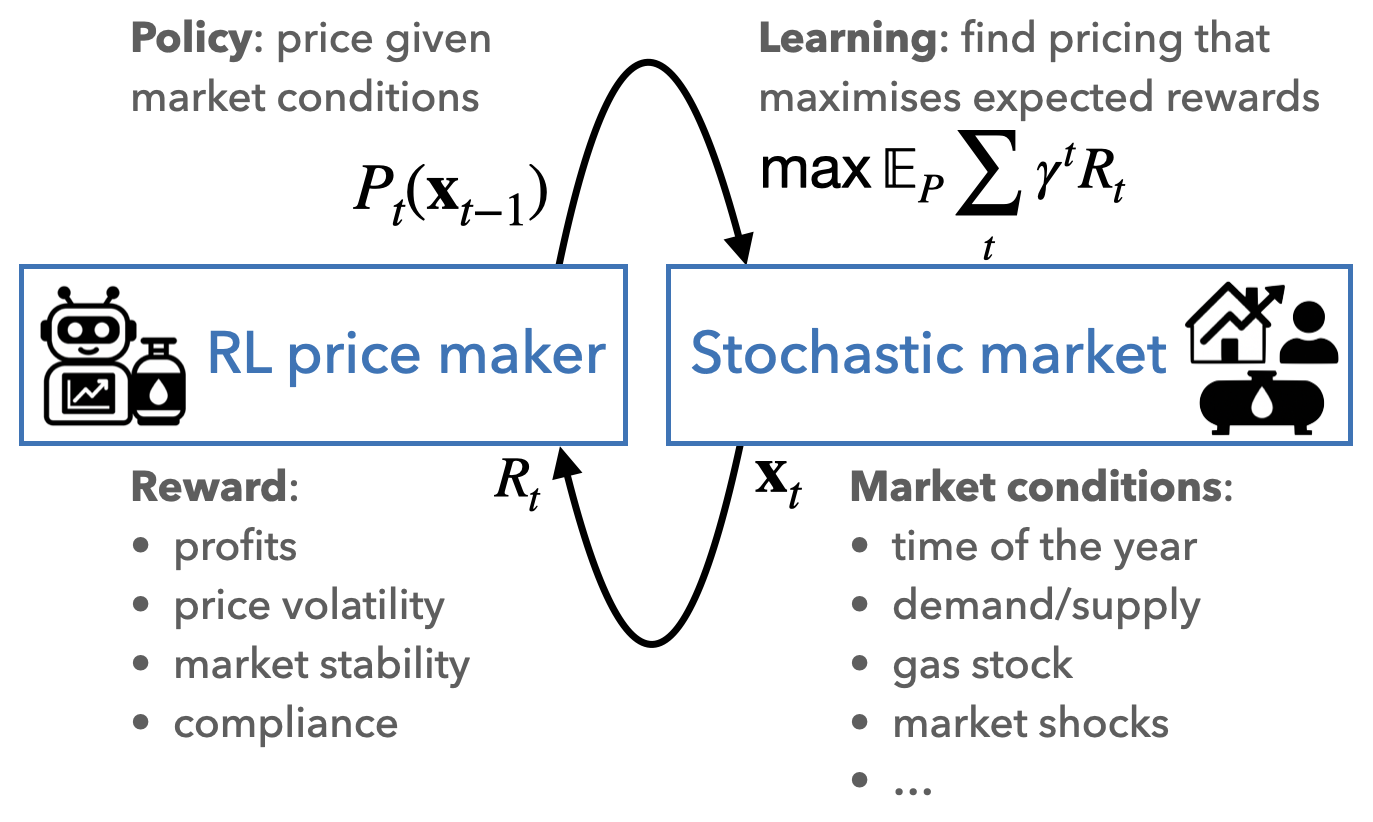}
    \caption{Illustration of the \algoName simulator.
    \normalfont
    The RL agent learns a policy $P_t(\mathbf{x}_{t-1})$ for the price of the natural gas at time $t$ given the market conditions at time $t-1$.
    The policy is learned via the maximisation of the expected value of discounted future rewards through repeated interactions with a stochastic market simulator.
    The instantaneous reward $R_t$ of the RL agent increases for increasing profits, but it decreases for increasing price volatility, lack of market clearing and non-compliance with regulations.
    The instantaneous vector of market conditions $\mathbf{x}_{t}$ includes signals such as the time of the year and the current values of demand and supply, stock of gas, and market shocks.
    }
    \label{fig:illustration}
\end{figure}

\section{The \algoName simulator}
\label{sec:simulator}

Our \algoName simulator is composed of two integrated components.
The first component is a carefully designed and calibrated market environment, which, given a price level as input, returns the quantities of gas demanded and supplied at that level.
The environment reproduces stylised facts such as the stickiness of demand and supply, seasonal variation in demand, and the persistence of stochastic shocks.
The second component is a reinforcement learning (RL) agent - the storage operator - which sets gas prices and uses its storage facilities to fill the imbalances between demand and supply that are generated by its pricing policy. The agent has multiple objectives, which are embedded in its reward structure: 1) ensuring market clearing by never running out of stored gas or storage capacity; 2) maximising profits; 3) minimising price volatility, consistently with the public-private nature of the storage operator; 4) fulfilling any refilling mandates imposed by the government.
We model the storage operator as a price setter in agreement with standard microeconomic practice for agents with market power. 
In real markets, large storage operators can influence prices at the margin by timing injections and withdrawals and by adjusting the size and composition of their trades, thereby shifting the net balance of supply and demand. 
The model assumes that the operator recognises this influence and acts strategically, internalising how its pricing policy affects market clearing and future states of the system. 
Under these conditions—market power plus strategic behaviour—it is natural to represent the operator's policy as a price rule rather than a pure quantity rule: the agent selects a price, and the environment maps that price into quantities demanded and supplied, with the storage facility bridging any imbalance subject to inventory constraints. 
This framing does not imply that the operator literally sets the market price unilaterally or that other market participants are passive; rather, it is a reduced-form representation equivalent to a monopolist choosing along a perceived residual demand curve. 
Modelling the price directly as the policy variable simplifies the interface between the agent and the market environment and provides a transparent way to encode objectives related to volatility and refilling mandates while preserving the economic content of strategic market power.

In Sec \ref{subsec:rl_env} we describe the gas market environment, while in Sec \ref{subsec:rl_agent} we describe the RL storage-operator agent with its observations, actions and rewards. 
An illustration of the simulator, highlighting the two components, is provided in Figure \ref{fig:illustration}.

\subsection{The RL environment}
\label{subsec:rl_env}
Time is discrete, and a unitary time increment represents a month. 
At each time $t$, the environment starts by including a given price $P_t$ into demand and supply log signals ($p^d$ and $p^s$ respectively) via the following functions
\begin{align}
\begin{split}
p^d_t &= \ln\bigl[\lambda_d\,e^{p^d_{t-1}} + (1 - \lambda_d) P_t], \\
p^s_t &= \ln\bigl[\lambda_s\,e^{p^s_{t-1}} + (1 - \lambda_s) P_t].
\end{split}
\end{align}
The exponentially weighted moving average of past signals allows for a ``sticky'' evolution of demand and supply.
As highlighted in \cite{emiliozzi2025unveiling}, the full impact of a change in the spot market price of gas on supply and demand is not realised immediately, but it unfolds over time as agents gradually adjust their behaviour (e.g., by switching to less gas-intensive technologies when prices rise).

\begin{table}[t]
\centering
\begin{tabular}{@{}lll@{}}
\toprule
\textbf{Symbol} & \textbf{Description} & \textbf{Value} \\
\midrule
% N            & Months per year        & 12    \\
$T$   & Total episodic steps (30 years) & 360   \\
\addlinespace
$\eta_d$     & Demand elasticity      & 0.20  \\
$\lambda_d$  & Demand stickiness      & 0.975 \\
$\rho_d$     & Demand AR(1) persistence  & 0.98  \\
$\sigma_d$   & Demand shock volatility       & 0.01  \\
$\mathcal{K}$ & Demand Fourier components     & $\{1, 2, 3, 4, 6\}$  \\
\addlinespace
$\eta_s$     & Supply elasticity  & 0.30  \\
$\lambda_s$  & Supply stickiness  & 0.95  \\
$\rho_s$     & Supply AR(1) persistence  & 0.75  \\
$\sigma_s$   & Supply shock volatility  & 0.04  \\
\addlinespace
$I_{\max}$   & Maximum capacity   & 3.0   \\
$\tau$       & Monthly storage cost  & 0.005 \\
$r$          & Monthly interest rate    & 0.0025\\
\addlinespace
\midrule
\addlinespace
L            & Action lower bound        & 0.01  \\
U            & Action upper bound        & 100.0 \\
$\gamma$ & Discount factor & 0.99 \\
\addlinespace
$\theta_v$ & Price volatility penalty & 20 \\
$\theta_m$      & Market-clearing penalty   & 1000 \\
$\theta_n$    & Annual threshold penalty & 750 \\
\bottomrule
\end{tabular}
\caption{Environment parameters (top) and RL agent parameters (bottom) of the \algoName simulator.}
\label{tab:all_params}
\end{table}

\begin{figure*}[t]
  \centering
  \includegraphics[width=0.98\textwidth]{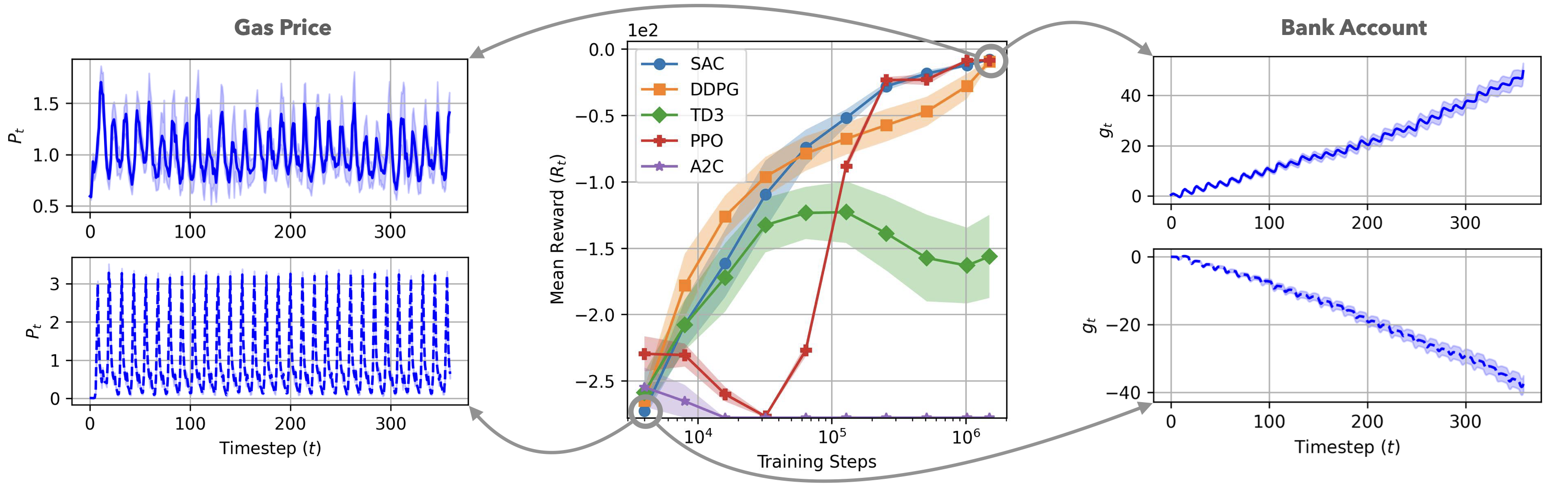}
  \caption{SAC outperforms other RL schemes for \algoName and yields realistic-looking time series.
    \normalfont
    The \textbf{centre} panel shows the mean cumulative episodic test rewards of five standard RL schemes as a function of the number of training steps.
    %
    % Mean and standard errors are computed using 10 independent training runs. \aldo{check: 10 runs?}
    %
    SAC stands out as the best-performing RL scheme, achieving better rewards more reliably than its competitors.
    The other panels show the mean trajectories of the price ($P_t$, \textbf{left} panels) and bank account ($g_t$, \textbf{right} panels) as learned by the SAC agent at 4000 steps (bottom rows) and at 1.5 million steps (top rows).
    The RL agent very quickly learns to set prices according to the season, as apparent by the periodicity of the 4000-steps pricing trajectory, but this is not sufficient to achieve good profits, as indicated by the 4000-steps bank account trajectory.
    However, at the end of training, the RL agent learns a much more sophisticated pricing policy that is able to achieve good profitability.
    }
  \label{fig:learning}
\end{figure*}

The log price signals computed as above determine, in turn, the log-demand ($d_t$) and log-supply ($s_t$) as
\begin{align}
\begin{split}
d_t &= S_t - \eta_d p^d_t + u^d_{t}   \\
s_t &= \eta_s p^s_t + u^s_{t},
\label{eq:demand-supply}
\end{split}
\end{align}
where $\eta_d$ and  $\eta_s$ are price-elasticity parameters, $S_t$ is a component that captures the seasonality of the demand, and $u^d_t$ and $u^s_t$ are stochastic demand and supply shifters.
The seasonal demand component is a truncated Fourier series
\begin{equation}
    S_t = \sum_{k \in \mathcal{K}} \left[ a_k \, \cos\left( \phi_t \,  k \right) + b_k \, \sin\left( \phi_t \, k \right) \right],
    \label{eq:fourier}
\end{equation}
where $\phi_t = 2\pi t / 12$.
The coefficients $a_k$ and $b_k$ are estimated from the monthly time series of gas consumption in Italy.
In Eq.~\eqref{eq:fourier}, the set $\mathcal{K}$ should include integer divisors of 12, so that the seasonal component has yearly periodicity.
The stochastic shifters $u_{d,t}$ and $u_{s,t}$ follow the AR(1) processes
\begin{align}
\begin{split}
u^d_t &= \rho_d \, u^d_{t-1} + \sigma_d \, \epsilon_{d,t} \\
u^s_t &= \rho_s \, u^s_{t-1} + \sigma_s \, \epsilon_{s,t} ,
\end{split}
\end{align}
where $\rho_d$ and $\rho_s$ are the autoregressive coefficients, $\sigma_d$ and $\sigma_s$ are volatilities, and $\epsilon^d_{t}$ and $\epsilon^s_{t}$ are i.i.d. standard normal random variables.

Note that the above reduced-form specification of demand and supply refers to a single national gas market—the Italian market in our simulations—but it does not necessarily imply that the market is perfectly isolated from other markets, as both demand and supply may include components stemming from linkages with other markets (e.g., via pipeline or Liquid-Natural-Gas facilities).

Once the log-demand $d_t$ and log-supply $s_t$ signals are evaluated, the excess demand is defined as
\begin{equation}
    D_t = e^{d_t} - e^{s_t},
\end{equation}

The excess demand is the amount of natural gas that needs to be withdrawn from storage (or injected into it if negative). 
Denote the amount of natural gas in storage at time $t$ by $I_t$ and the maximum storage capacity by $I_{\max}$. 
Then, the new inventory will be
\begin{equation}
I_{t+1}=
\begin{cases}
I_t - D_t, &   -(I_{\max}-I_t)\le D_t \le I_t\\[1pt]
0, &   D_t>I_t \,\,\,\,\,\,\,\,\,\,\,\,\,\,\,\,\,\,\,\,\,\,\,\,\,\, \text{ [unmet demand]}\\[1pt]
I_{\max}, &  D_t< -(I_{\max}-I_t)  \text{ [wasted supply]}
\label{eq:unmet_demand_wasted_supply}
\end{cases}
\end{equation}
The second and third conditions in the above equation both imply a market failure.
Specifically, a market failure occurs either when there is a strong excess of supply ($D_t< -(I_{\max}-I_t)$), resulting in a desired accumulation of natural gas above capacity and ultimately in a waste of gas, or when there is a large excess of demand ($D_t > I_t $), above the available inventories and hence impossible to meet.
Market failures, which the RL agent will learn to avoid because they are associated to steep negative rewards, are tracked by an indicator variable $m_t$ equal to one if a market failure occurs, and equal to zero otherwise.

The net amount of gas bought by the storage operator at time $t$ is given by 
\begin{equation}
    \Delta I_t = I_t - I_{t-1},
\end{equation}
with a negative value indicating that gas has been sold.

Selling or acquiring gas changes the bank account of the storage operator $g_t$, which in general evolves as
\begin{equation}
g_t = (1+r) \,  g_{t-1}
- \tau \, I_{t-1}
- P_t \Delta I_t 
+ \mathds{1}_{\{t=T \}} I_t P_t^m 
% + P_{mean} I_t. % TODO this is not true in general!
\label{eq:bank_account_evolution}
\end{equation}
In Eq.~\eqref{eq:bank_account_evolution}, 
the first term is the interest rate on bank deposits and 
$\tau$ is a proportional storage cost. 
The term $P_t \Delta I_t$ is the change in bank deposits due to selling or acquiring natural gas at the price $P_t$.
Finally, at the end of the simulation, the agent sells its residual stock of natural gas $I_t$ at the liquidation price $P_t^m = (e^{p_t^d} + e^{p_t^s} ) /2$.

\subsection{The RL agent}
\label{subsec:rl_agent}

\vspace{0.2cm}
\noindent
\textbf{Action space.}
The agent's action at each time step $t$ is a single continuous scalar $p_t$, which is then exponentiated to obtain the market price $P_t = e^{p_t}$.
For numerical stability reasons, we clip the action in a wide but bounded range of values
$p_t \in [\ln L,\ln U]$.
By choosing $p_t$, the agent implicitly determines the supply and the demand for gas in that month, and hence the net flow of gas into storage (injection or withdrawal).

\vspace{0.2cm}
\noindent
\textbf{State space.}
The state space $\mathbf{x}_t$ provides a full representation of market conditions at time $t$.
It is a real‑valued vector with nine components, namely
\begin{equation}
    \mathbf{x}_t =(S_t,\cos \left( \phi_t \right),\sin \left(\phi_t \right),u^d_t,u^s_t,p^d_{t},p^s_{t},\ln(0.5+I_{t}), p_{t}),
\end{equation}
where $S_t$ is the seasonal component of demand,
the cosine and sine of $\phi_t = 2\pi t / 12$ are simple signals that identify the month of the year, and the stockpiles of natural gas are provided in the form $\ln(0.5+I_t)$ for numerical reasons (to roughly keep them in a symmetric range around zero).
More details on each of these components are provided in Sec.~\ref{subsec:rl_env}.

\begin{figure*}[htbp]
  \centering
\includegraphics[width=0.195\linewidth, trim={0.75cm 0 0.35cm 0}, clip]{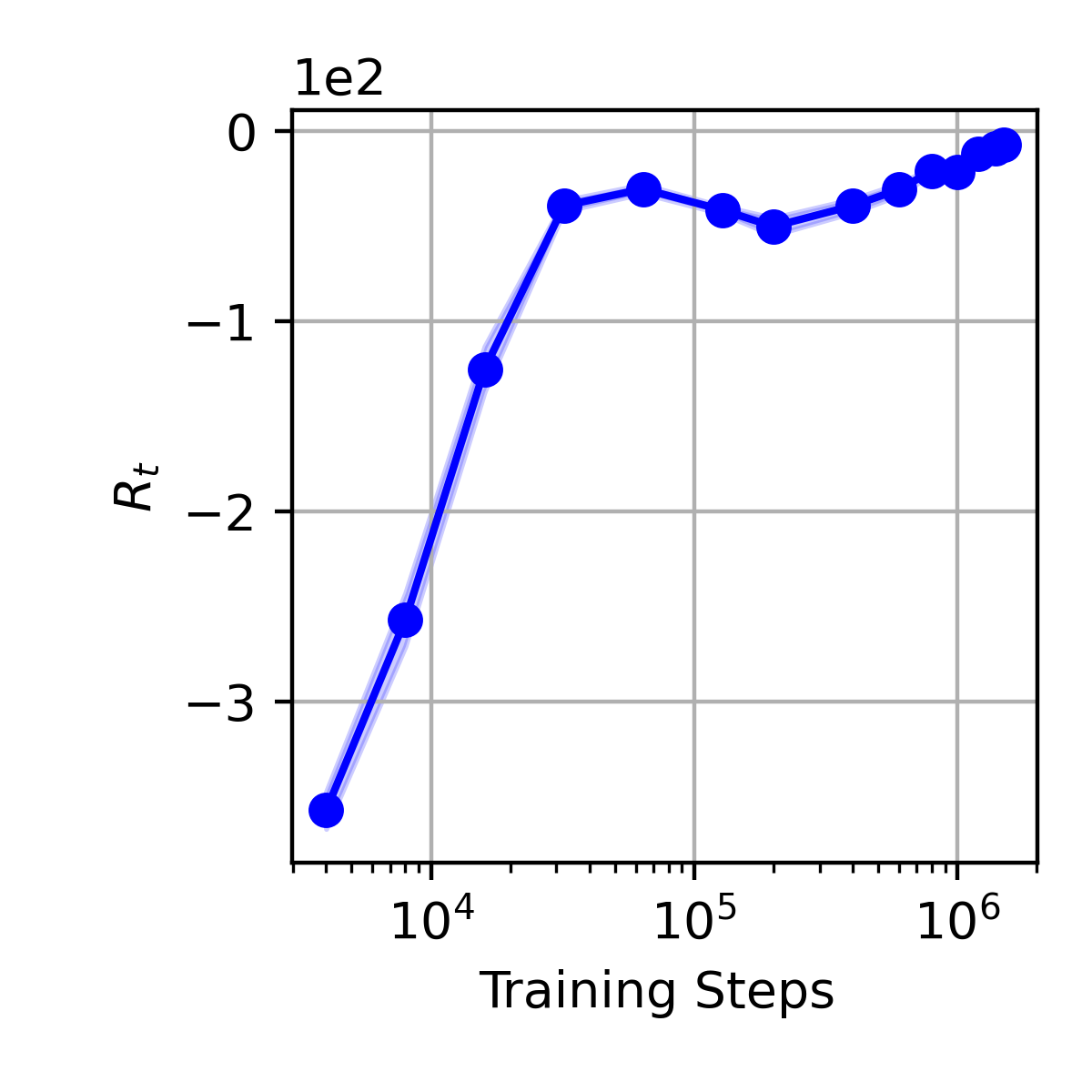}
    \includegraphics[width=0.195\linewidth, trim={0.35cm 0 0.35cm 0}, clip]{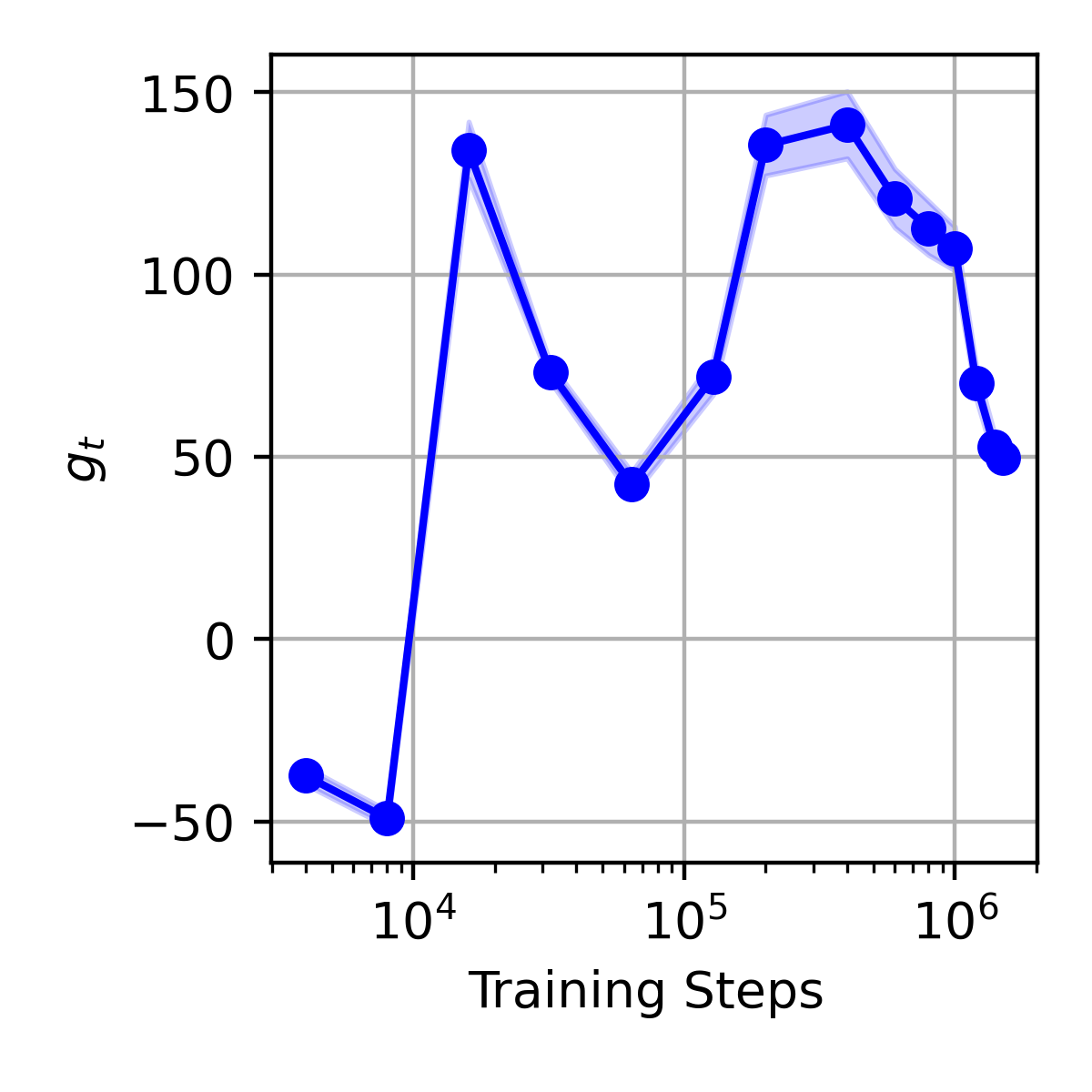}
    \includegraphics[width=0.196\linewidth, trim={0.35cm 0 0.35cm 0}, clip]{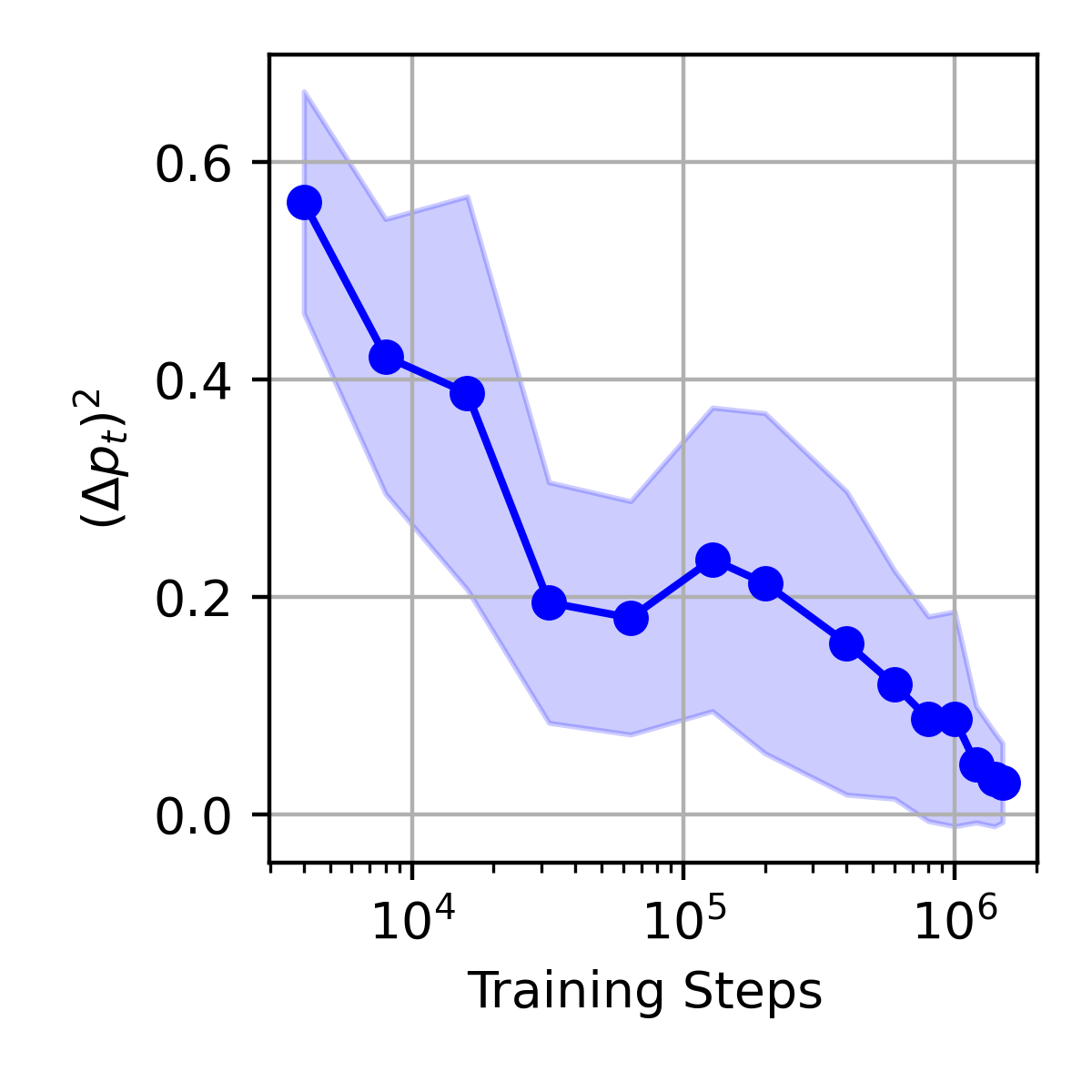}
    \includegraphics[width=0.196\linewidth, trim={0.35cm 0 0.35cm 0}, clip]{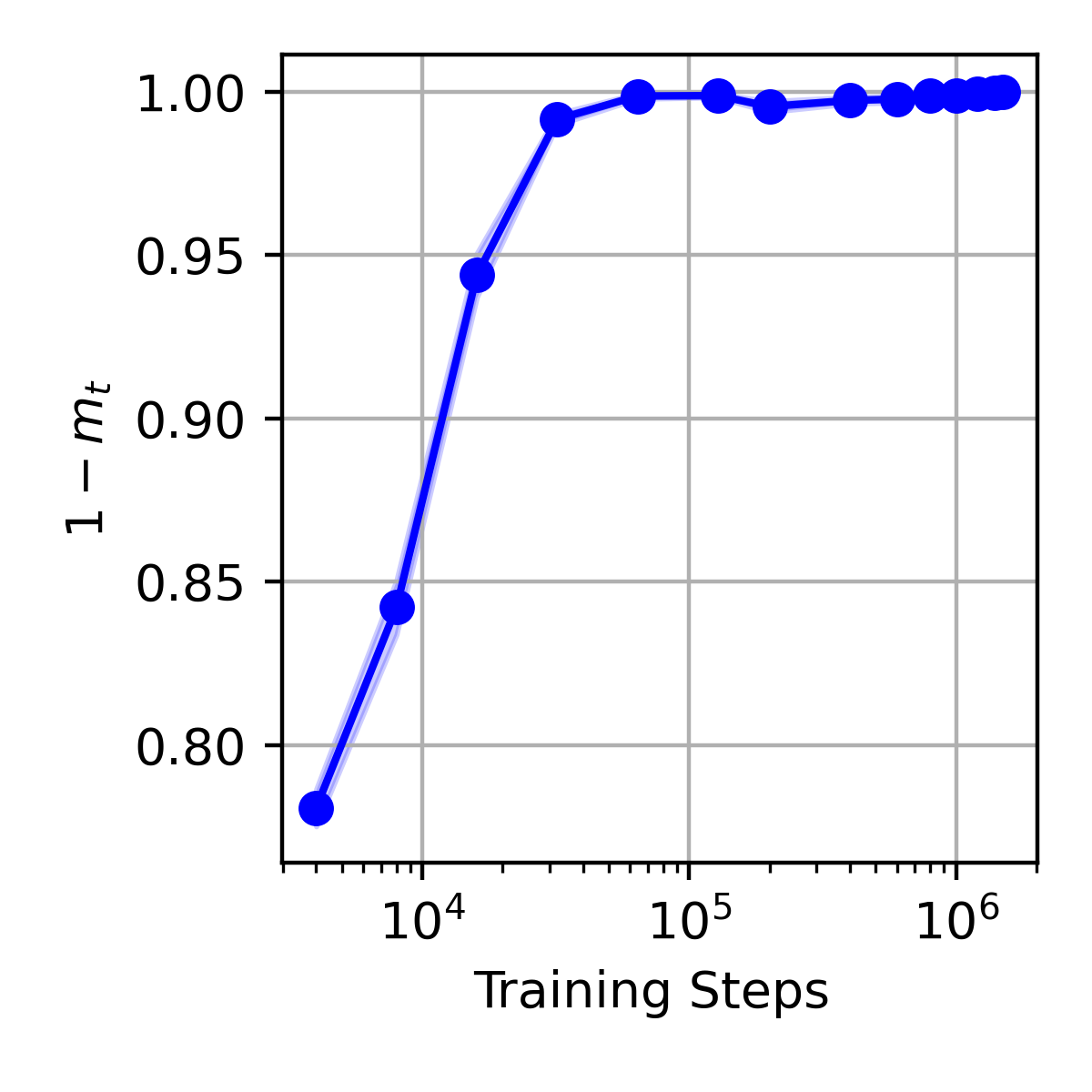}
    \includegraphics[width=0.198\linewidth, trim={0.25cm 0 0.35cm 0}, clip]{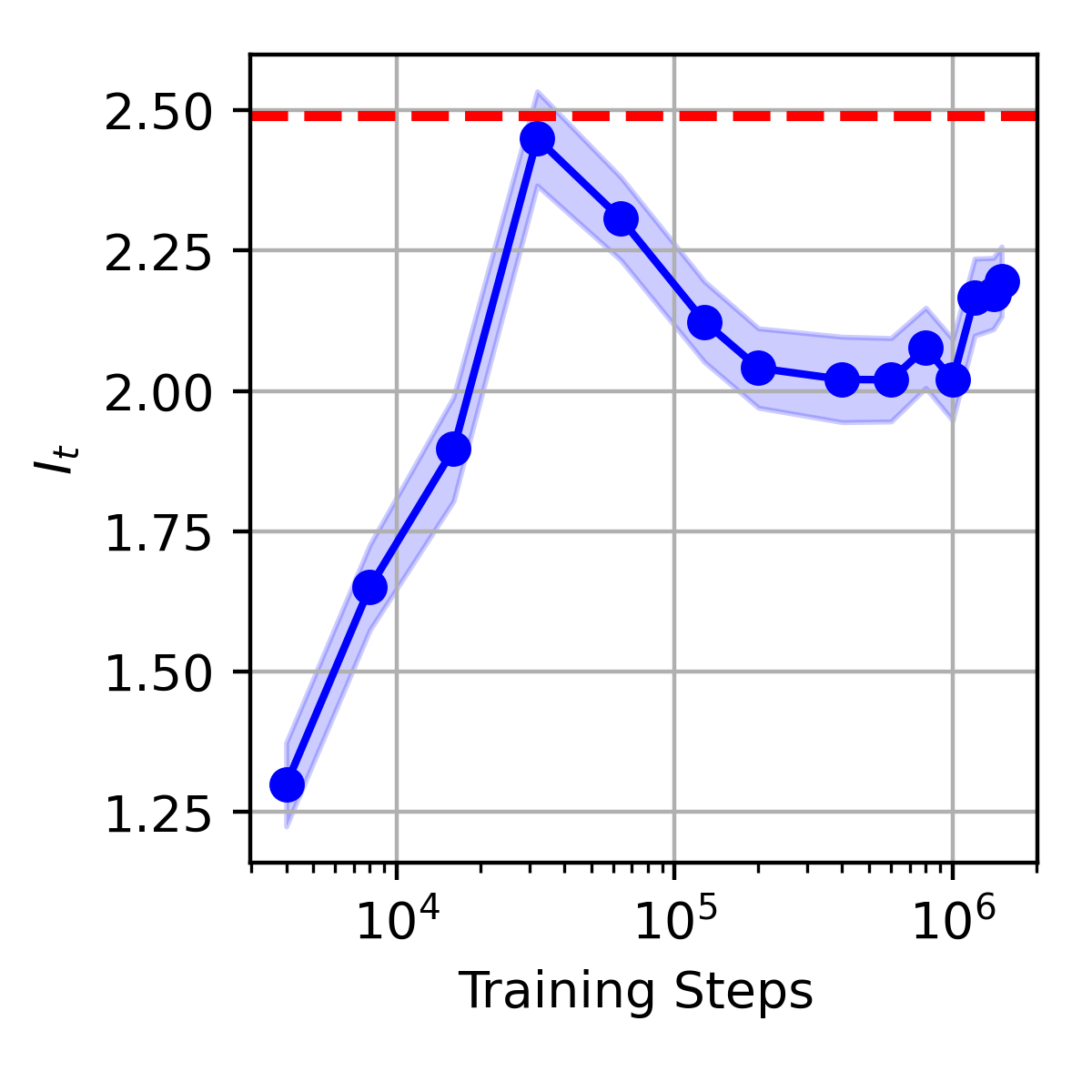}
  \caption{
  \algoName yields profitability, stable markets and reasonable stockpiles.
  \normalfont
  The figure illustrates how the model's test performance, evaluated using different metrics, changes as the number of training steps increases. 
  All panels show the mean and the 95\% confidence intervals on the mean computed with 50 repetitions, for the best SAC model saved at different checkpoints. 
  Specifically, from left to right the different panels present the means of: reward ($R_t$), bank account ($g_t$), price volatility ($(\Delta p_t)^2$), market success rate ($1-m_t$), and the level of inventories ($I_{t}$) at the beginning of November.
  With increased training, the reward rises until it converges; the bank account increases with more steps with an uneven progression, as sometimes profitability is lost in favour of a lower volatility.
  The market‑success metric levels off much earlier, at around 32.000 learning steps.
  The inventories in November rise up to 2.5 (or 83\% of the storage capacity) at the beginning of training before settling around 2.2 (73\% of the storage capacity).
  }
  \label{fig:metrics_vs_training_steps}
\end{figure*}

\vspace{0.2cm}
\noindent
\textbf{Reward.}
The RL agent learns a deterministic policy $P(\mathbf{x})$ to set the price $P_t$ given the market conditions at a previous time $\mathbf{x}_{t-1}$.
The policy is learned by maximising $\mathcal{R}$, the expected sum of discounted rewards under $P(\mathbf{x})$
\begin{equation}
\mathcal{R} = \mathbb{E}_{P} \left[ \sum_{t=1}^{T} \gamma^t R_t \right].
\end{equation}
The reward $R_t$ at time $t$ is defined as
\begin{equation}
R_t = \Delta g_t - \theta_{v} (\Delta p_t)^2 - \theta_{m} m_t (1 + \tilde{m}_t)- \theta_{n} n_t (1 + \tilde{n}_t)
\label{eq:reward}
\end{equation}
where
\begin{itemize}
  \item $\Delta g_t = g_t - g_{t-1}$ denotes the change in the bank account. By aiming to increase this term over time, the agent is effectively maximising its profits.
  \item $(\Delta p_t)^2 = (p_t - p_{t-1})^2$ is the per-period contribution to price volatility and $\theta_{v}$ is a positive scalar. The gas storage operator is incentivised to design pricing policies that balance profit maximisation with market stability. This trade-off reflects the operator's dual private–public nature.
  \item $m_t$ was defined before as a categorical variable equal to $1$ when the market does not clear and to $0$ otherwise.
  $\tilde{m}_t$ quantifies the severity of the market failure, being $\tilde{m}_t = D_t - I_t$ in case of unmet demand and $\tilde{m}_t = | D_t | - (I_{\max}-I_t)$ in case of wasted supply (see Eq. \eqref{eq:unmet_demand_wasted_supply}).
  By setting the positive scalar $\theta_{m}$ equal to a sufficiently high value, the agent can be incentivised to find pricing policies that avoid market failures (almost) always.
  \item $n_t$ is a categorical variable equal to $1$ if the inventory is below a threshold in a given month of the year (November in our simulations), and $0$ otherwise. 
  $\tilde{n}_t$ quantifies the amount by which the minimum storage threshold is not met. For instance, with an 83\% threshold one would have $\tilde{n}_t = 0.83 \, I_{\max} - I_t$.
  This last term is used to model government-mandated minimum storage requirements.
\end{itemize}

\section{Experimental setup}
\label{sec:experimental_setup}

\vspace{0.2cm}
\noindent
\textbf{Model parameters.}
The parameters of the \algoName environment are given in the top part of Table \ref{tab:all_params}.
They are calibrated to the Italian gas market to ensure that the simulations yield realistic values for: 1) the volatilities and dynamic elasticities of demand and supply, as estimated in \cite{emiliozzi2025unveiling}; 2) the ratio between storage capacity and average monthly gas consumption; 3) the proportion between monthly storage costs and gas prices; 4) the seasonal variation in demand.

The parameters shaping the rewards of the RL agent are given in the bottom part of Table \ref{tab:all_params}.
In particular the parameters $\theta_v$, $\theta_m$ and $\theta_n$ that determine the trade-offs among the agents' multiple objectives, are calibrated with the goals of guaranteeing that: 1) the volatility of gas prices determined endogenously in the model matches its real-world counterpart; 2) the market clearing and refilling constraints are (almost) always met.

\vspace{0.2cm}
\noindent
\textbf{Training and testing.}
We implemented \algoName using well-known open source libraries in Python. 
Specifically, we implemented the \algoName environment following the standard interface offered by the Gymnasium package \cite{towers2024gymnasium}, which allowed us to use RL algorithms directly from Stable-Baselines3 \cite{raffin2021stable}.
In our experiments, we consider the following RL algorithms for the gas-storage agent: Deep Deterministic Policy Gradient (DDPG) \cite{LillicrapHPHETS15}, Twin Delayed Deep Deterministic policy gradient (TD3) \cite{fujimoto2018addressing}, Advantage Actor-Critic (A2C) \cite{mnih2016asynchronous}, Soft Actor Critic (SAC) \cite{haarnoja2018soft}, Proximal Policy Optimization (PPO) \cite{schulman2017proximal}.
For each algorithm, we perform five independent training runs of 1.5 million steps.
Moreover, while we rely on default hyperparameter choices in our baseline training runs, we check the robustness of our results to changes in learning rates, model checkpointing and selection strategies, and actor-critic network architectures (number of hidden layers and neurons per hidden layer).

To compare the performance of the different algorithms we proceed as follows.
We perform 10 training runs with different seeds for each model. 
Then, for each seed, we perform 5 test runs and compute a mean reward for each seed.
Finally, for each algorithm, we report a mean value and a standard error by averaging the mean values associated to its seeds.

% we perform 5 test runs for each of the 10 trained models and plot the average value and standard errors only across the different training runs.
% \marco{Da chiarire meglio. Cosa sono i 10 trained models?}
%
To analyse the performance of the simulator, we select the single algorithm reaching the highest sum of rewards at the end of training.
Using such a model, unless otherwise stated, we perform 50 test runs and plot the average values and standard errors of the different variables over these test runs.

\vspace{0.2cm}
\noindent
\textbf{Reproducibility.}
The code for the \algoName environment used to perform our experiments is available in open source at \url{https://anonymous.4open.science/r/GasRL-8DD6}. A reimplementation of \algoName in Julia is also available at \url{https://github.com/aldoglielmo/GasRL.jl}.
% \aldo{TODO: check this before submitting}

\section{Results}
\label{sec:results}

\subsection*{SAC outperforms all other learning schemes.}
In the central panel of Figure \ref{fig:learning}, we report the mean returns achieved by the different RL schemes considered as a function of the number of training steps.
While SAC, DDPG and PPO achieve very similar rewards after 1.5M training steps, SAC outperforms all other schemes in terms of learning stability.
In this respect, TD3 
and PPO clearly underperform with respect to SAC.
TD3 also exhibits much larger reward fluctuations around the mean as compared to the other methods, as highlighted by the much larger error bands.
DDPG is able to challenge the performance of SAC up to around 50k steps of training, before a performance deterioration.
Finally, the A2C algorithm, possibly due to a high sensitivity to hyperparameters, is found to be incapable of learning using the standard parameterisation. 

Given its excellent stability and performance on the \algoName environment, we use the SAC algorithm to carry out the bulk of our empirical analysis. 
The results shown in the rest of the paper are obtained from a SAC agent trained until convergence.
We decided to rely on the hyperparameter choices proposed as defaults in the Stable-Baselines3 package \cite{raffin2021stable} after observing that the performance of trained agents does not change significantly by increasing the number of hidden layers (to 3) and/or the number of neurons in those layers (by 2x or 4x), and/or by reducing the learning rate (by 3x or 10x).

\begin{figure}
    \centering
    \includegraphics[width=1.0\linewidth]{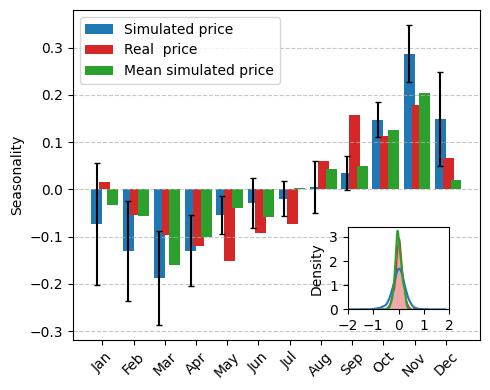}
    \caption{\algoName price volatilities and seasonality are consistent with real-world data.
    \normalfont
    The main panel depicts the seasonality of natural-gas prices as computed on real-world data (red bars) and on synthetic data generated by the \algoName simulator (blue bars).
    Given the high variability in the seasonality of simulated data, we also show the seasonality computed on the prices as averaged over multiple runs (green bars).
    The inset in the bottom right shows kernel density estimates of the distribution of the first log differences, for the same three series and using the same colour code.
    In both graphs, the coherence between the real-world data and the output of the \algoName simulator is clear.
    %    
    % \aldo{ [TIZIANO: put the final figure of the right size (e.g., 4.5x3.5) and the inset of the right size (e.g., 2x2). The inset figure should ideally be composed directly in Python. ]}
    % 
    }
    \label{fig:volatility_seasonality}
\end{figure}

\subsection*{\algoName yields realistic-looking time series.}

The left and right panels of Figure \ref{fig:learning} show mean and standard errors of price trajectories (left) and bank account trajectories (right) for a poorly trained agent (bottom) and a fully trained agent (top).
It is interesting to note that the RL agent quickly learns the necessity of adjusting the natural gas price to the seasonality of the demand function, as evidenced by the large periodic oscillations of the price $P_t$ shown in the bottom left panel.
However, this basic cyclical strategy is not sufficient for the agent to become profitable over time, as evidenced by the downward slope of the bank account $g_t$ shown in the bottom right panel.
The fully trained RL agent exhibits a much more nuanced and sophisticated price policy, as shown in the top left panel.
The price oscillations here have much smaller variability, they are much less affected by seasonality, and are instead much more responsive to current market shocks.
Overall, they appear much more realistic than the poorly trained alternatives.
This sophisticated policy succeeds at making the gas-storage operator profitable, as evidenced by the upward sloping curve in the top right panel.

\begin{figure*}[t]
  \centering
    \includegraphics[width=0.247\linewidth]{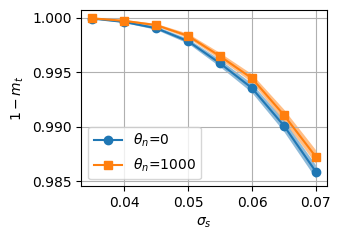}
    \includegraphics[width=0.247\linewidth]{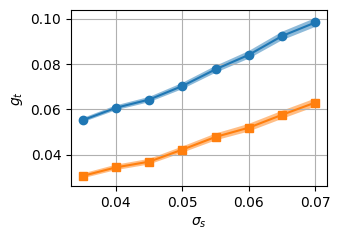}
    \includegraphics[width=0.247\linewidth]{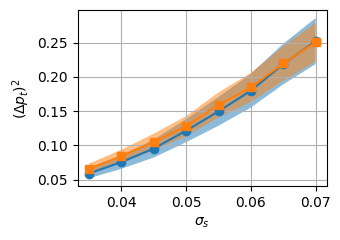}
    \includegraphics[width=0.247\linewidth]{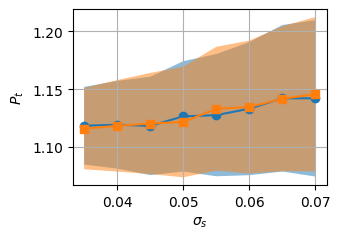}
    
  \caption{\algoName suggests that a regulatory threshold on gas stockpiles can increase market stability.
  {\normalfont The figure illustrates test results for different supply-shock volatility test-values $\sigma_s$ (i.e., what happens when the storage operator unexpectedly faces a volatility of supply shocks that is different from the one used to optimise the policy). 
  Each panel shows the mean and the 95\% confidence interval around the mean, computed with 1000 repetitions. 
  From left to right, the panels report the mean values of the market success rate ($1-m_t$), the bank account ($g_t$), the price volatility level ($(\Delta p_t)^2$) and the price level ($P_t$), as a function of supply shock volatility ($\sigma_s$), for the baseline model ($\theta_n = 0$, blue circles) and the regulated model ($\theta_n = 1000$, orange squares).
  Introducing a penalty for not reaching the 83\% minimum-storage threshold seems to slightly improve market success robustness ($1-m_t$). 
  However, this comes at the expense of reduced profits for the gas storage operator, as evidenced by significantly lower bank account values ($g_t$), and at the cost of a slightly increased price volatility $(\Delta p_t)^2$.
  Interestingly, the average price level ($P_t$) is roughly unaltered by the regulatory requirement.}
  }
  \label{fig:robustness}
\end{figure*}

\subsection*{\algoName yields profitability, stable markets and reasonable stockpiles.}

The results shown in Figure \ref{fig:metrics_vs_training_steps} demonstrate the effectiveness of the RL agent's learning phase.
Specifically, we note from the first panel that the agent's mean reward ($R_t$) computed at the end of the 30-year test horizon steadily increases by increasing the number of training steps.
The growth of the bank account ($g_t$) is less steady; nonetheless, it exhibits a substantial improvement by the end of the training phase compared to its outset.
In parallel, the price volatility ($(\Delta p_t)^2$) steadily decreases as shown in the third panel.
Notably, the fourth panel clearly shows how market failures are virtually eliminated over a 30-year test horizon.
This behaviour aligns with real-world imperatives: the gas storage operator needs to be profitable, but it also needs to maintain a low price volatility given its public-private nature, and it needs to prioritise the avoidance of market disruptions given the high costs such events impose.
The last panel shows the gas inventories ($I_t$) stocked by the agent in November, the month in which recent EU regulation enforces a re-filling threshold.
Interestingly, even without any compliance enforcement ($\theta_n$ is zero in these runs), the November inventories naturally reach the 83\% threshold at the beginning of training, before settling at around 2.2 (or 73\% of the storage capacity).

\subsection*{\algoName price volatility and seasonality are consistent with real-world data.}

In Figure \ref{fig:volatility_seasonality} we compare the seasonality and volatility of the price series generated by our \algoName simulator with those found in real-world data (TTF front-month gas futures prices recorded at the end of each month in the period 2010-2024). 
We compute approximate percentage changes by taking the first differences of log-prices. 
Then, we compute price seasonality by running linear regressions of price changes on monthly dummies, that is, months one-hot-encoded as independent variables. Finally, we report the estimated regression coefficients, which measure the seasonal component of the price. This method is sometimes called `deterministic seasonal model'  \cite{ABEYSINGHE1994259}.
We perform these computations on historical real-world data (the TTF future prices; shown in red), on the 50 simulated time series (in blue), and on a single time series that is obtained by averaging the 50 simulated price trajectories (in green).
The second methodology used to produce simulated data (averaging) should give rise to more precise estimates of the seasonal coefficients, as it smooths out simulation-to-simulation variability.
With both methodologies, the seasonal patterns found in simulated data closely mirror those observed in historical data, aside from the fact that seasonal troughs occur slightly earlier in simulations than in reality. 
For both real and simulated series, the peak value is recorded in November, coinciding with the final deadline for reaching the minimum inventory-refilling level.
This alignment underscores the model's effectiveness at faithfully reproducing real-world price dynamics.

%Regarding volatility, the simulations exhibit comparable values, 26\% for the determinist and 21\% for the alternative setup, versus a real‐world volatility of just under 20\%, calculated on the historical TTF price series from 2010 to the present.\cite{ABEYSINGHE1994259}.
%
% \aldo{TODO: ask Marco Taboga about this}

The figure also shows kernel density estimates of the distributions of real and simulated data. 
The standard deviation of the simulated log-price difference is 27\%, higher than the historical one, calculated on the whole 2010-2024 sample (17\%), but close to that experienced in recent years (25\% in the 2020-2024 period). 
This might reflect the fact that some of the model parameters, such as demand and supply persistence, were calibrated on more recent data.

\subsection*{\algoName suggests that a storage threshold can improve resilience to supply shocks.}
Finally, we demonstrate the use of the \algoName simulator by analysing the effects of introducing a mandatory gas storage threshold. 
Specifically, we set a minimum refilling level of 83\% of total capacity to be reached by the beginning of November. 
This value was proposed during recent negotiations to revise the EU regulation on gas storage and appeared likely to be enacted into legislation at the time we conducted our simulations.
For this exercise, we proceed as follows.
We activate the last term in the reward function (Eq.\eqref{eq:reward}) if the natural gas inventories $I_t$ do not reach $83\%$ of the maximum capacity $I_{\max}$ at the beginning of November, and compare the `baseline' model (trained with $\theta_n=0$) with a `regulated' model (trained with $\theta_n=1000$).
We train both models exclusively on the original value of the supply shock volatility, $\sigma_s=0.04$, as given in Table \ref{tab:all_params}, and check how the two respond to increasing supply volatilities up to $\sigma_s=0.07$, or 75\% more than the original training value.
The results of this experiment are shown in Figure \ref{fig:robustness}.

The first panel clearly shows that the introduction of the regulatory constraint on gas stockpiles gives rise to markets that are more resilient to increases in supply shock volatility, as measured by the average market success $1-m_t$.
The improvement is small, yet statistically significant, and it is larger for larger supply shock volatilities $\sigma_s$.
However, this increased market robustness comes with two costs.
First, the regulated RL gas operator achieves positive, yet smaller profitability since the bank account ($g_t$) stabilises at lower levels compared to the baseline scenario.
Second, the regulated model is forced to sacrifice some price stability as evidenced by the larger price volatility ($(\Delta_{p_t})^2$). 
The increase in price volatility is very small, but appears significant for low values of supply shock volatilities.
Interestingly, the regulatory constraint appears to have no measurable effect on the average price level ($P_t$).

\section{Conclusions}
\label{sec:conclusions}

This study introduces \algoName, a simulator that couples a calibrated stochastic representation of the Italian natural-gas market with a monopolistic storage operator modelled by deep reinforcement learning (RL).
We showcase how \algoName can be used for both market analysis and regulatory design. 
The environment definition, parameter calibration and code base are publicly released in open source for full reproducibility and to facilitate extensions.

We benchmark five state-of-the-art algorithms, finding that the Soft-Actor-Critic scheme is superior to its competitors, robustly achieving high rewards with smaller training fluctuations.
Once trained, the SAC agent generates realistic-looking pricing trajectories that increase profits monotonically, eliminate all market failures, keep price volatility within empirically plausible bounds, and lead to reasonable stockpiles of natural gas.
Furthermore, the simulator reproduces key stylised facts of the Italian market.
The seasonality coefficients of synthetic prices match well with those estimated from historical data, and the distribution of first-difference log-returns exhibits a spread that is compatible with its real-world counterpart.
This realism arises endogenously as no price series was used in training, a fact that corroborates the ability of the RL gas storage operator to learn economically coherent behaviours from the interaction with the calibrated market environment.
Leveraging \algoName, we explored the EU debate on mandatory storage levels, finding that imposing an 83\% November threshold can lead to a small yet significant improvement in market resilience to adverse supply shocks.
This comes at the cost of lower profitability and slightly lower price stability, but has no effect on the overall price level.

With \algoName, we combined economic modelling with modern RL schemes to obtain a powerful tool that allows for flexible market and policy analysis.
Several directions could be pursued to extend the present work. 
First, the computational efficiency of the \algoName simulator could be significantly improved by enabling GPU-accelerated training, for instance by adopting frameworks such as rlax \cite{deepmind2020jax} or directly Jax \cite{jax2018github}.
This would greatly facilitate more extensive hyperparameter tuning and allow for a finer calibration of the environment to real-world data. 
Second, while this study focused primarily on the simulator architecture and its baseline performance, it can be interesting to perform a more in-depth exploration of the policy implications derived from different regulatory scenarios.
Third, future work could assess the robustness of the learned policies with respect to alternative specifications of the environment, including different demand or supply shock processes, elasticity estimates, or institutional constraints. 
Lastly, the introduction of additional agents—such as competing storage operators or traders—could open the way to multi-agent extensions of \algoName, allowing for a richer analysis of market dynamics and strategic behaviour. 
In particular, the current version of the model treats the Italian market as a closed system, in which interactions with foreign markets can be captured only by the reduced-form supply and demand equations. 
Multi-agent variants could explicitly address the international dimension of the natural-gas market, which—like many other commodity markets—is partly segmented at the national level, due to limited infrastructure connecting it with other markets, and partly open to international trade thanks to some cross-border pipeline connectivity and Liquid-Natural-Gas facilities.

%\begin{acks}
%
%\aldo{We would like to thank [MARCO: DO WE WANT TO THANK ANYONE HERE?]
%
%The views and opinions expressed in this paper are those of the
%authors and do not necessarily reflect the official policy or position
%of Banca d’Italia.}
%
%\end{acks}

\section*{Code availability}
In the interest of reproducibility, the code for the \algoName is available in open source. The original Python version, used for the experiments, is available at \url{https://github.com/TizianoBacaloni/GasRL}, a reimplementation in Julia is available at \url{https://github.com/aldoglielmo/GasRL.jl}.

\bibliographystyle{ACM-Reference-Format}

% \newpage

\bibliography{references}

\end{document}